\def\year{2020}\relax
\documentclass[letterpaper]{article} 
\usepackage{aaai20}  
\usepackage{times}  
\usepackage{helvet} 
\usepackage{courier}  
\usepackage[hyphens]{url}  
\usepackage{graphicx} 
\usepackage{graphicx}  

\usepackage{graphicx}
\usepackage{algorithmicx}  
\usepackage{algpseudocode}  
\usepackage{amsmath}  
\usepackage{color,xcolor}
\usepackage{pifont}
\usepackage{bm}
\usepackage{amsfonts}

\usepackage[normalem]{ulem}
\usepackage{multirow}
\usepackage{amsmath}
\usepackage{url}

\usepackage{amssymb}
\usepackage{amsopn}
\usepackage{subcaption}
\usepackage{graphicx}
\usepackage{multirow}
\usepackage[english]{babel}
\usepackage{bm}
\usepackage{array}
\usepackage{booktabs}
\usepackage{color}
\usepackage[textsize=scriptsize]{todonotes}

\title{Neural Simile Recognition with Cyclic Multitask Learning
	\\and Local Attention}
\author{
Jiali Zeng\textsuperscript{\rm 1}\thanks{Equal contribution, work done when the first author was an intern at Tencent, Beijing, China.}, 
\ Linfeng Song\textsuperscript{\rm 2}$^{*}$, 
\ Jinsong Su\textsuperscript{\rm 1}\thanks{Corresponding author.}, 
\ Jun Xie\textsuperscript{\rm 3}, 
\ Wei Song\textsuperscript{\rm 4}, 
\ Jiebo Luo\textsuperscript{5} \\ 
\textsuperscript{\rm 1}Xiamen University, Xiamen, China \ \ \
\textsuperscript{\rm 2}TencentAI Lab, Bellevue, USA \\
\textsuperscript{\rm 3}Mobile Internet Group, Tencent Technology Co., Ltd, Beijing, China \\ 
\textsuperscript{\rm 4}Capital Normal University, Beijing, China \ \ \
\textsuperscript{\rm 5}University of Rochester, Rochester NY, USA \\
lemon@stu.xmu.edu.cn, \ \ freesunshine0316@gmail.com, \ \ jssu@xmu.edu.cn \\
stiffxie@tencent.com, \ \ wsong@cnu.edu.cn, \ \ jluo@cs.rochester.edu \\
}
 
\begin{document}

\maketitle

\begin{abstract}
Simile recognition is to detect simile sentences and to extract simile components, i.e., {\it tenors} and {\it vehicles}.
It involves two subtasks: {\it simile sentence classification} and {\it simile component extraction}.  
Recent work has shown that standard multitask learning is effective for Chinese simile recognition, but it is still uncertain whether the mutual effects between the subtasks have been well captured by simple parameter sharing.
We propose a novel cyclic multitask learning framework for neural simile recognition, which stacks the subtasks and makes them into a loop by connecting the last to the first.
It iteratively performs each subtask, taking the outputs of the previous subtask as additional inputs to the current one, so that the interdependence between the subtasks can be better explored.
Extensive experiments show that our framework significantly outperforms the current state-of-the-art model and our carefully designed baselines, and the gains are still remarkable using BERT.
Source Code of this paper are available on https://github.com/DeepLearnXMU/Cyclic.
\end{abstract}

\section{Introduction}
\begin{figure}[!ht]
	\centering
	\includegraphics[width=1.0\linewidth]{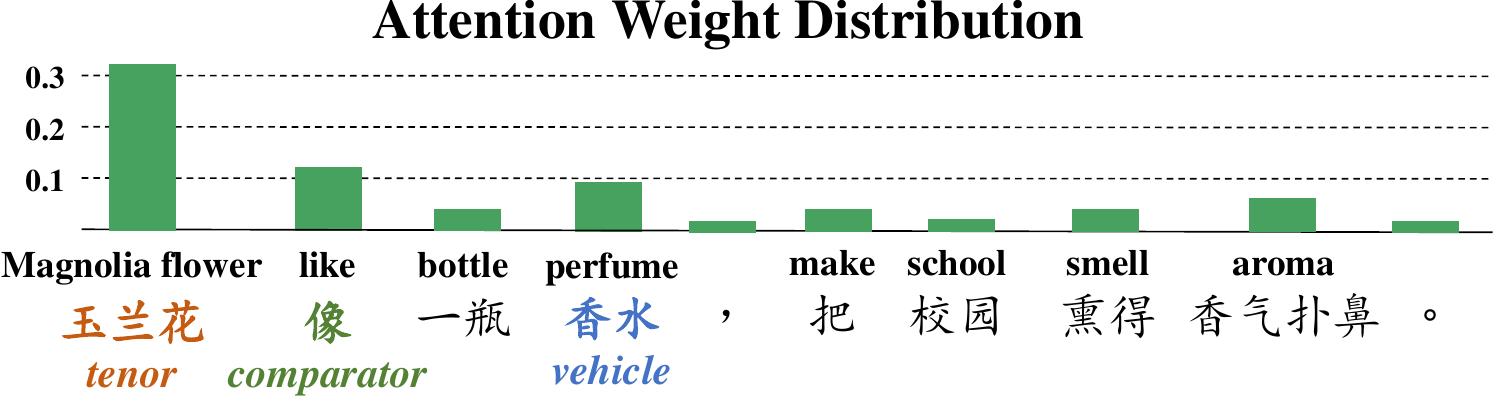}
	\caption{\label{Exam_intro}
		Attention weights generated by the simile sentence classifier of the conventional multitask learning \cite{Liu:EMNLP2018},  showing that the simile sentence classifier tends to focus more on simile components.
	}
\end{figure}

Simile is a special type of metaphor that compares two objects (called {\it tenor} and {\it vehicle}) of different categories using comparator words 
such as ``{\it like}'', ``{\it as}'' or ``{\it than}''. 
A Chinese simile sentence is shown in Figure \ref{Exam_intro}, where the tenor ``{\it Magnolia flower}'' and the vehicle ``{\it perfume}'' are compared using comparator ``{\it like}''. 
Typically, simile recognition involves two subtasks \cite{Liu:EMNLP2018}: {\it Simile Sentence Classification}, which discriminates whether a sentence containing a comparator is a simile sentence, and {\it Simile Component Extraction}, which aims to extract the tenor and the vehicle in a simile sentence, respectively. 

It is of great importance to study simile.
Simile recognition is potentially beneficial for NLP applications, such as sentiment analysis (e.g. hate speech detection), dialogue understanding and question answering, because users sometimes use simile to express their emotions.
In addition, simile recognition can help language learners to better understand the implicit meanings expressed by simile in books and novels by highlighting simile components.
However, simile recognition is very challenging, with one reason being that simile sentences have very similar syntactic and semantic structures to the normal sentences, hindering the feasibility of standard NLU techniques, such as syntactic and semantic parsing.
Even though comparator words can provide some hints, they are also frequently used in literal comparisons, which introduces great ambiguity to this task.

Previous approaches of simile recognition are primarily based on handcrafted linguistic features and syntactic patterns \cite{Li:CIP2008,Niculae-Yaneva:ACL2013,Niculae:JSSP2013,Niculae-Mizil:EMNLP2014}, which are 
inefficient on new languages and domains due to the extra time for feature engineering.
Inspired by the successful applications of neural multitask learning on many NLP tasks \cite{Liu:IJCAI2016,Zhang:ACL2016,Luo:EMNLP2015,Miwa:ACL2016}, 
\citeauthor{Liu:EMNLP2018} (2018) investigated a standard multitask learning framework on simile recognition, which significantly outperforms the existing methods. 
Specifically, they apply a bi-directional LSTM \cite{Hochreiter:NC1997} to encode the representations of each input sentence, and then the encoding results are taken as features shared by an attention-based sentence classifier, a CRF-based component extractor \cite{Lafferty:ICML2001} and a language model that serves as an auxiliary task for additional supervision signals.

Despite their success, the multitask learning framework of \citeauthor{Liu:EMNLP2018} (2018) suffers from two major drawbacks. 
\textbf{First}, simple parameter sharing is unable to fully exploit the semantic interdependence between the two subtasks of simile recognition.
It is intuitive that the results of these subtasks can be beneficial to each other. 
The potential simile components usually get higher attention weights during simile sentence classification. 
Taking Figure \ref{Exam_intro} as an example, 
the attention weights for tenor ``{\it Magnolia flower}'' and vehicle ``{\it perfume}'' are much higher than those of the other words.
Therefore, simile component extractor can be more precise with the 
information about potential tenor and vehicle 
(attention distribution)
from the simile sentence classifier. 
Moreover, it will be easier for simile sentence classification if we have identified the tenor and vehicle through component extraction, since they directly determine whether the sentence is a simile.

\textbf{Second}, both tenors and vehicles are usually close to comparators.
In a standard simile benchmark \cite{Liu:EMNLP2018}, the average distances from tenors to comparators and from vehicles to comparators are 3.0 and 4.3, respectively, while the average sentence length is 29.5.
As a result, the global attention mechanism used by \citeauthor{Liu:EMNLP2018} (2018) can suffer from attention errors, since it considers all words in a sentence.
Back to Figure \ref{Exam_intro}, irrelevant words, such as the words after ``{\it ,}'', distract the global attention model and own attention weights significantly larger than zero.

\begin{figure}[!t]
\centering
\includegraphics[width=0.95\linewidth]{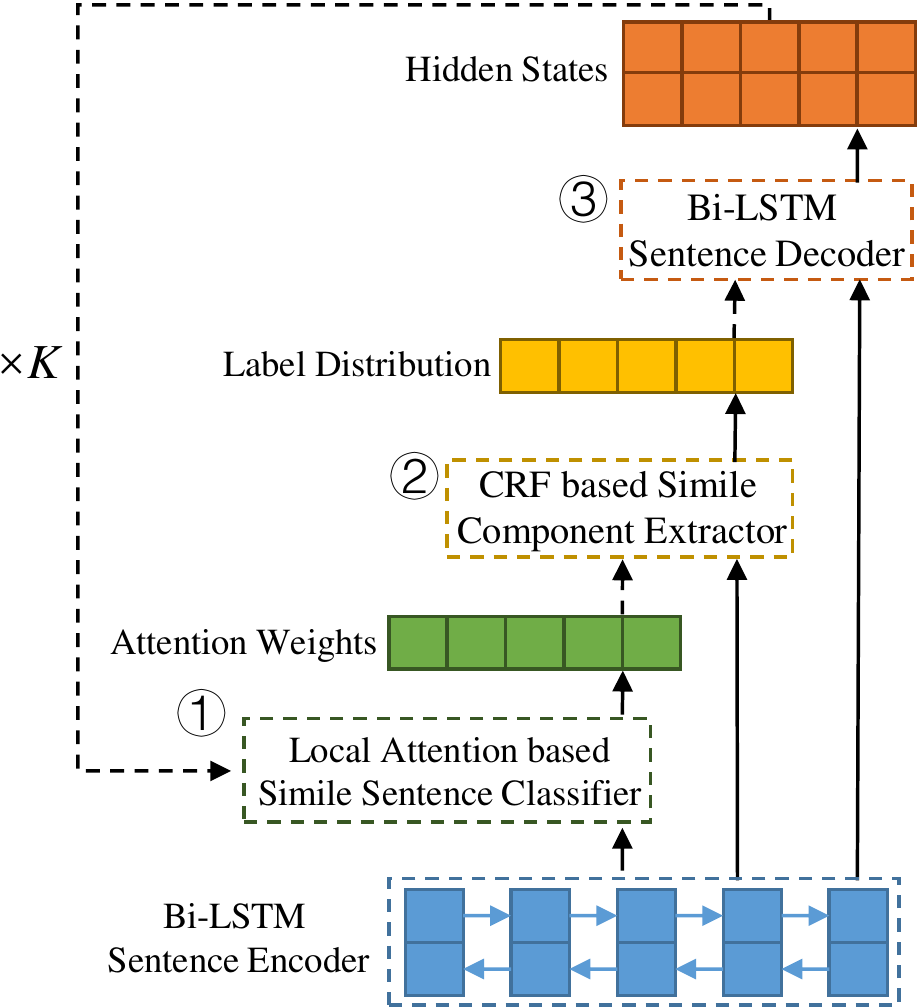}
\caption{
\label{OurModel}
The architecture of our framework. Following previous work \cite{Liu:EMNLP2018}, \ding{194} is taken as an auxiliary sub-task for additional supervision.
}
\end{figure}

To overcome the above drawbacks, we propose a novel cyclic multitask learning framework with local attention for neural simile recognition. 
Figure \ref{OurModel} shows our framework, which captures the correlations of its subtasks by feeding the output of each subtask into the next.
It organizes the subtasks as a cycle that is executed for $K$ times, thus all subtasks further benefit from all others.

Taking $K=1$ as an example,
{\bf first}, 
a Bi-LSTM encodes the input sentence and produces a sequence of word representations. 
{\bf Then},  
a local attention model is applied to the local sequence of word representations around the comparator word, and the induced local context vector is fed into a simile sentence classifier.
{\bf Next}, we concatenate the attention weights generated by the local attention and word representations, before sending the results to a CRF layer to extract simile components via sequence labeling.
{\bf Afterwards}, the label distribution and the word representations are concatenated as the input of the sentence decoder to reconstruct the original sentence.
{\bf Finally}, the decoder states are summed with the word representations, and the results are the input for the following simile classification.

Overall, our contributions are three folds:
\begin{itemize}
	\item We propose a novel cyclic multitask learning framework for neural simile recognition. Comparing with standard multi-task learning, this framework better models the inter-correlation among its sub-tasks.
	\item We introduce a local attention mechanism for simile sentence classification. To our knowledge, no previous work has explored local attention on this task.
	\item Our framework shows superior performance over carefully designed baselines with or without pretrained BERT \cite{Devlin:NAACL2019}, introducing the new state-of-the-art performance in the literature.
\end{itemize}

\section{Our Framework}
In this section, we give a detailed description of our proposed framework. 
As shown in Figure \ref{OurModel}, our cyclic framework concatenates a local attention based simile sentence classifier (\ding{192}), a CRF based simile component extractor (\ding{193}), and a Bi-LSTM sentence decoder (\ding{194}) as a cycle, where the output of each module is fed as additional input to its successor module. In this way, the interdependence between different subtasks can be better exploited in our framework. 
In addition, it contains a Bi-LSTM sentence encoder that provides shared features to these modules (\ding{192}, \ding{193} and \ding{194}).
Note that our framework executes for $K$ times, and the execution path is \ding{192}$\rightarrow$\ding{193}$\rightarrow$\ding{194}$\rightarrow$\ding{192} when $K=1$.

\subsection{Bi-LSTM based Sentence Encoder}\label{encoder}
Given an input sentence $X=(x_1,x_2,\dots,x_N)$,
we follow Liu et al.~\shortcite{Liu:EMNLP2018} to first map its words to embeddings. 
Then, a Bi-LSTM is applied to produce a sequence of word representations $H=(h_1,h_2,\dots,h_N)$ that are shared by our subtasks.
The forward LSTM reads the sentence from left to right to learn the representation of each word $x_i$ as $\overrightarrow{h}_{i}$.
Similarly, 
the backward LSTM reversely scans the source sentence and learns each representation $\overleftarrow{h}_{i}$. 
Finally, for each word $x_i$,
the representations from two LSTMs are concatenated to form the word representation $h_i=[\overrightarrow{h}_{i}$,$\overleftarrow{h}_{i}]$.

\subsection{Local Attention based Simile Sentence Classifier}
\begin{figure}[!t]
	\centering
	\includegraphics[width=1.0\linewidth]{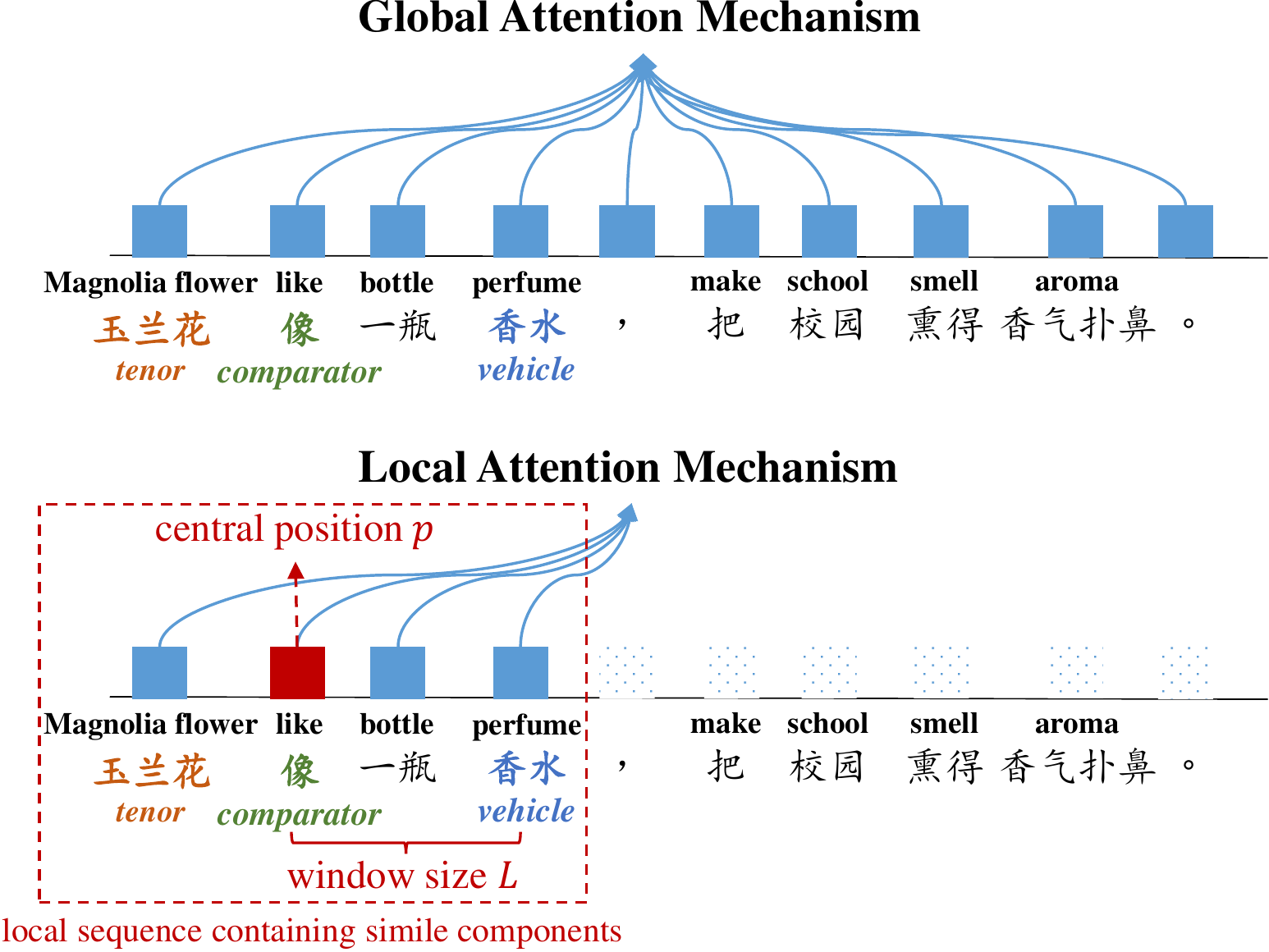}
	\caption{\label{LocalAtt}
		Global attention mechanism and our proposed local attention mechanism.
	}
\end{figure}
Simile sentence classification is a binary classification task determining whether a sentence contains any simile or not. 
As analyzed previously, the adopted global attention mechanism of \citeauthor{Liu:EMNLP2018} (2018) considers all words.
Hence, it is not suitable for the simile sentence classification, which mainly depends on the tenor and the vehicle around the comparator.
To address this issue, 
we base our simile sentence classifier on a \emph{local} attention mechanism.
Compared with a global attention mechanism, our local attention mechanism only focuses on a dynamically choosed local context surrounding the comparator word.

We contrast our local attention mechanism with the global attention mechanism in Figure \ref{LocalAtt}. 
For more details, 
since the comparator word, such as "like", "than" or "as", is given in each sentence, we first choose the position of the comparator word as the central position $p$, and then dynamically generate a context window size $L$ as follows:
\begin{align} \label{eq:l}
L &= \frac{N}{2} \cdot \sigma (\emph{v}^{T}_{w} tanh(\sum_{i=1}^{N} W_{o} h_i + W_{p}h_p)),
\end{align}
where $h_p$ is the representation vector of the central word, $W_p$, $W_o$, and $\emph{v}_w$ are learnable parameters.
As the next step, we perform an attention operation on the local sequence of word representations \{$h_{p-L},..., h_p, ..., h_{p+L}$\} to generate a local context vector $r$ via
\begin{equation}
\begin{aligned}
r&=\sum_{i=p-L}^{p+L} \lambda_{i} h_i, \\ \label{eq4}
\lambda_{i} &= softmax(\emph{v}^{T}_{a}tanh(W_{o} h_i)),
\end{aligned}
\end{equation}
where $W_o$ has been defined in Eq.(1) and $\emph{v}_{a}$ is a model parameter.
Next, we stack a feed-forward network on $r$ to induce high-level features. Finally, these features are used as inputs of a $softmax$ layer to conduct classification:
\begin{equation}\label{q5}
p(y|H) = softmax(W^{0}_c {\rm ReLU}(W^{1}_{c}r)),
\end{equation}
where $W^{0}_c$ and $W^{1}_c$ are model parameters\footnote{Right now we follow previous work to assume there being one or zero simile for each instance. For future multi-simile extension, we simply sum all $r$s before Equation \ref{q5}.}.

As shown in Figure \ref{OurModel}, there are two cases for simile sentence classification within our cyclic framework: for the first case, only word representations $H$ are available for producing the local attention weights as additional input to simile component extractor, while for the other case, both $H$ and the states $S$=($s_1, s_2, ..., s_N$) of the Bi-LSTM sentence decoder (\ding{194} in Figure \ref{OurModel}) are available. We directly take the sum of $H$ and $S$ as $H$, i.e., $H=H+S$, to conduct simile classification. To make a unified definition, the loss function for simile sentence classification is defined as:
\begin{equation}
\mathcal{J}_{sc} = -log(p(y|H,S)).
\end{equation}
Intuitively, $S$ contains useful information of simile component extractor and sentence decoder, which can be directly propagated to simile sentence classifier by incorporating $S$ into it.

\subsection{CRF based Component Extractor} \label{ce}
We follow Liu et al.~\shortcite{Liu:EMNLP2018} to implement simile component extraction as a sequence labeling task. 
As mentioned before, the results of simile sentence classification can be beneficial to simile component extraction. 
Thus, we augment the word representations $H$ with the attention weights generated from Equation \ref{eq4}: 
\begin{equation}
\begin{aligned}
{\widetilde H} &=({\widetilde h_1}, ..., {\widetilde h_N}), \\
{\widetilde h_i} &=\left\{
\begin{aligned}
& [h_i, \lambda_i] & i \in \{p-L,...,p+L\} \\
& [h_i, 0] & else
\end{aligned}
\right.,
\end{aligned}
\end{equation}
and then we stack a CRF layer on ${\widetilde H}$. Formally, the score of a predicted sequence $Y^{'}$=($y_1, y_2, ..., y_N$), where $y_i \in D$ and $|D|=d$, is defined as
\begin{equation}
\varphi ({\widetilde H},Y^{'})=\sum^{N}_{i=1} A_{y_{i-1},y_i} + \sum^{N}_{i=1}M_{i,y_i},
\end{equation}
where $A \in \mathbb{R}^{d \times d}$  is a transition matrix updated during training, and $A_{y_{i-1},y_i}$ records the transition score from label $y_{i-1}$ to $y_{i}$; similarly,
$M$=($m_1$,...,$m_N$) $\in \mathbb{R}^{N \times d}$ is the emission matrix and $M_{i,y_i}$ indicates the score of assigning tag $y_i$ to $x_i$.
Specifically, $m_i$ is a $d$-dimensional label distribution vector generated by feeding ${\widetilde h_i}$ into a single-layer network with activation function $tanh$ and a $softmax$ layer.
Finally, the probability of the label sequence $Y$ given the sentence $X$ is
\begin{align}
&p(Y|{\widetilde H}) = \frac{exp(\varphi ({\widetilde H},Y))}{\sum_{\tilde{Y}} exp(\varphi({\widetilde H},\tilde{Y}))}. 
\end{align}
To train this extractor, we minimize the standard log-likelihood loss function:
\begin{align}
\mathcal{J}_{ce} &= -log(p(Y|{\widetilde H}) \notag \\
&= -\varphi({\widetilde H},Y) + log \sum_{\tilde{Y}} e^{\varphi({\widetilde H},\tilde{Y})}.
\end{align}

\subsection{Bi-LSTM based Sentence Decoder}
Due to the small number of available training instances,
we follow Liu et al.~\shortcite{Liu:EMNLP2018} to incorporate language modeling into our cyclic framework as an auxiliary task, which can help Bi-LSTM encoder better model the sentence information.

For more details,
we concatenate each label distribution vector $m_i$ from component extractor with word representation $h_i$ to produce the initial state $s_0$ of the sentence decoder:
\begin{equation}
\begin{aligned}
{\overline h_i} &= [h_i,m_i], \\
s_0 &= W^{0}_d\sum^{N}_{i=0} {\overline h_i},
\end{aligned}
\end{equation}
where $W^0_d$ is a learnable matrix.
Next, 
the forward LSTM takes the previous hidden state $\overrightarrow{s}_{t-1}$ and the previous word embedding $e(x_{t-1})$ as input to produce the hidden state $s_t$ at the $t$-th timestep:
\begin{equation}
\overrightarrow{s}_{t} = \text{LSTM}(e(x_{t-1}), \overrightarrow{s}_{t-1}),
\end{equation}
and then predict the current word $x_t$ in the following way:
\begin{equation}
\begin{aligned}
&p(x_{t}|x_{<t})=softmax(\overrightarrow{W}^{1}_d \overrightarrow{q}), \\
&\overrightarrow{q} = tanh(\overrightarrow{W}^{2}_d\overrightarrow{s}_{t} + \overrightarrow{W}^{3}_d e(x_{t-1})),
\end{aligned}
\end{equation}
where $\overrightarrow{W}^{*}_d$, $*$$\in$[1,2,3] are trainable matrix parameters.
Formally, the loss function for the forward sentence decoder is defined as
\begin{equation}
\overrightarrow{\mathcal{J}}_{fd}=-\sum^{N}_{t=1} log(p(x_{t}|x_{<t})).
\end{equation}
Similarly, the backward decoder is the same as the forward decoder, but with different model parameters. 
Equations are omitted for space limitation. The backward loss function is defined as
\begin{equation}
\overleftarrow{\mathcal{J}}_{bd}=-\sum^{N}_{t=1} log(p(x_{t}|x_{>t})).
\end{equation}
Finally, the loss function for the whole decoder is defined as the sum of those in two directions:
\begin{equation}
\mathcal{J}_{sd}=\overrightarrow{\mathcal{J}}_{fd} + \overleftarrow{\mathcal{J}}_{bd}.
\end{equation}

\subsection{Overall Training Objective}
The final training objective over an instance, which contains a sentence, a simile tag, and a sequence of component labels, becomes
\begin{equation}
\mathcal{J} = \alpha \cdot \mathcal{J}_{sc} + \beta \cdot \mathcal{J}_{ce} + (1-\alpha-\beta) \cdot \mathcal{J}_{sd},
\end{equation}
where $\alpha$, $\beta$ (s.t. $\alpha$+$\beta<$ 1) are non-negtive weights assigned beforehand
to balance the importance among the three tasks.

\section{Experiments}
\subsection{Settings}
\subsubsection{Data}
We evaluate our model on a standard Chinese simile recognition benchmark \cite{Liu:EMNLP2018}, where each instance contains \emph{one} or \emph{zero} similes. 
Table \ref{tab:dataset} shows the basic statistics of this dataset.
We follow \citeauthor{Liu:EMNLP2018} (2018) to conduct 5-fold cross validation: the dataset is first equally divided into 5 folds.
For each time, 4 folds are used as training and validation sets (80\% for training, 20\% for validation), and the remaining fold is used for testing.
For simile extraction, the components (tenors and vehicles) are tagged with the IOBES scheme \cite{Ratinov:CoNLL2009}. 
\begin{table}
\centering
\begin{tabular}{lc} 
\hline
\small
\#Sentence  & 11,337  \\
\#Simile Sentence  & 5,088   \\	
\#Literal Sentence  & 6,249   \\
\#Token  & 334K   \\
\#Tenor  & 5,183   \\
\#Vehicle & 5,119 \\
\#Unique tenor concept & 1,680 \\
\#Unique vehicle concept & 1,972 \\
\#Tenor-vehicle pair & 5,214 \\
\hline
\end{tabular}
\caption{Statistics of our simile dataset.}
\label{tab:dataset}
\vspace{-1.0em}
\end{table}

\subsubsection{Hyper-parameters}
For fair comparisons, we use the same hyper-parameters as \cite{Liu:EMNLP2018}.
In particular,
we use their pretrained 50-dimensional Word2Vec \cite{Mikolov:ICLR2013} embeddings, which are updated during training. 
For efficient training, we only use the sentences with at most 120 words.
The hidden sizes for Bi-LSTM encoder and decoder are 128.
The parameters between sentence encoder and bi-directional sentence decoder are shared.
Also, 
the word embeddings and the pre-softmax linear transformation in the sentence decoder are shared. 
The batch size is 80. The dropout rate is 0.5. We adopt {\it Adadelta} \cite{Zeiler:Arxiv2012} as the optimizer with a learning rate of 1.0 and {\it early stopping} \cite{Prechelt:NN1998}.
The optimal hyper-parameters $\alpha$=0.1, $\beta$=0.8 are chosen using the validation set.

\subsubsection{Contrast Models}
We compare the following baselines and models to study the effectiveness of our cyclic MTL:
\begin{itemize}
\setlength{\itemsep}{0pt}
\setlength{\parsep}{0pt}
\setlength{\parskip}{0pt}
	\item \textbf{ME \cite{Li:CIP2008}}.
	It is a maximum entropy model taking tokens, POS and dependency relation tags as features.
	\item \textbf{MTL \cite{Liu:EMNLP2018}}. A multitask learning framework, where the simile sentence classification, simile component extraction and sentence reconstruction are jointly modeled. It is the previous state-of-the-art system for simile sentence classification.
	\item \textbf{MTL-OP \cite{Liu:EMNLP2018}}. An ``Optimized Pipeline'' introduced by Liu et al.~\shortcite{Liu:EMNLP2018} for improving simile component extraction. 
	It involves two steps: it first uses 1-best results produced by a model jointly training their simile sentence classifier and language model to filter simile sentences; then, another model jointly training simile component extractor and language model is used to extract simile components from these sentences.
	Note that this model is just a pipeline for decoding.
	\item \textbf{MTL}. Our implemantation of {\it MTL \cite{Liu:EMNLP2018}}, where \emph{local} attention is adopted instead of a global one.
	\item \textbf{MTL-Pip}. It is a degraded variant of our framework without cyclic connection.
	Note that this is also novel as no previous work has investigated it on this task.
	\item \textbf{MTL-Cyc}. Our cyclic multitask learning framework.
\end{itemize}

\subsection{Effect of the executing number $K$}
\begin{table}
\centering
\small
\begin{tabular}{l|c|c}  
\hline
\textbf{{\it K}}  & \textbf{F1-score for \ding{192}} &  \textbf{F1-score for \ding{193}} \\
\hline
0  & 86.09  & 63.15  \\
1  & 86.62  & 73.33   \\	
2  & 86.65  & 73.40   \\
3  & 86.27  & 73.57   \\
4  & 85.89  & 72.83   \\
\hline
\end{tabular}
\caption{Experimental results on the validation set with different execution number $K$, where standard MTL results are shown when $K=0$.}
\label{tab:kresult}
\end{table}
Table \ref{tab:kresult} shows the validation results of our {\it MTL-Cyc} framework regarding the executing time $K$, where we show the results of {\it MTL} when $K=0$.
There are large improvements for both simile classification (task \ding{192}) and simile component recognition (task \ding{193}) when increasing $K$ from $0$ to $1$, showing the effectiveness of stacking the subtasks into a loop.
Further increasing $K$ from 1 to 2 only results in marginal improvements for both subtasks while introducing more running time, and their performances slightly go down when enlarging $K$ from 2 to 4.
All the evidence above indicates that our cyclic framework converges quickly, 
making it more practically useful.
Considering both efficiency and performance, we set $K=1$ for all experiments thereafter.

\subsection{Task 1: Simile Sentence Classification}\label{sec:task1}
\begin{table}
\centering
\small
\setlength{\tabcolsep}{1.4mm}{
\begin{tabular}{l|ccc}  
\hline
\textbf{Model}  & \textbf{Precision} & \textbf{Recall} & \textbf{F1-score} \\
\hline
ME \cite{Li:CIP2008}  & 76.61  & 78.32   & 77.45 \\
MTL \cite{Liu:EMNLP2018} & 80.84  & 92.20   & 86.15 \\
\hline
\ding{192}-Global \cite{Liu:EMNLP2018} & 77.51  & 88.95   & 82.84 \\
\ding{192}             & 79.76  & 88.25   & 83.79 \\
\hline
MTL(\ding{192}+\ding{193})          & 81.95  & 87.44   & 84.61 \\
MTL(\ding{192}+\ding{194})          & 81.45  & 88.96   & 85.04 \\
MTL-Pip(\ding{192}$\rightarrow$\ding{193})   & 81.72  & 89.67   & 85.51 \\
MTL-Pip(\ding{193}$\rightarrow$\ding{192})   & 81.50  & 89.26   & 85.20 \\
\hline
MTL(\ding{192}+\ding{193}+\ding{194})       & 81.60  & 92.10   & 86.53 \\
MTL-Pip(\ding{192}$\rightarrow$\ding{193}$\rightarrow$\ding{194}) & 81.39  &  \textbf{93.01}  & 86.81 \\
MTL-Pip(\ding{193}$\rightarrow$\ding{192}$\rightarrow$\ding{194}) & 81.80  & 91.99   & 86.59 \\
MTL-Cyc   & \textbf{82.12}  & 92.60  & \phantom{*}\textbf{87.04}* \\
\hline
\end{tabular}}
\caption{Main results on simile sentence classification. * indicates significant at $p<0.01$ over \emph{MTL(\ding{192}+\ding{193}+\ding{194})} with 1000 bootstrap tests \cite{efron1994introduction,koehn2004statistical}. For the remaining of this paper, we use the same measure for statistical significance.}
\label{tab:SC_result}
\end{table}

Table \ref{tab:SC_result} shows the experimental results on simile sentence classification. 
Overall, our \emph{MTL-Cyc} exhibits the best performance, outperforming the previous state of the arts: \emph{ME \cite{Li:CIP2008}} and \emph{MTL \cite{Liu:EMNLP2018}} and our baselines.
In addition, we have the following interesting observations:

\subsubsection{Effect of Local Attention} 
As shown in Table \ref{tab:SC_result} ({\bf Line 4-5}), when replacing the conventional global attention with our proposed local one, the performance of simile sentence classifier is improved by about 1 points.
This confirms the hypothesis that focusing on the local context of comparator is more suitable for detecting simile sentences.

\subsubsection{Effects of Simile Component Extraction and Sentence Reconstruction}
Here, we incrementally add simile component extraction and sentence reconstruction to explore their contributions to simile sentence classification under different frameworks: {\it MTL} and {\it MTL-Pip}.
From Table \ref{tab:SC_result} ({\bf Line 6-13}), we draw some conclusions:
\textbf{First}, 
when jointly modeling two subtasks, {\it MTL}(\ding{192}+\ding{193}), {\it MTL}(\ding{192}+\ding{194}), {\it MTL-Pip}(\ding{192}$\rightarrow$\ding{193}) and {\it MTL-Pip}(\ding{193}$\rightarrow$\ding{192}) all significantly outperform the single task model \ding{192}.
This indicates that there exists intense interdependence between the subtasks of simile recognition.
\textbf{Second},
both {\it MTL-Pip}(\ding{192}$\rightarrow$\ding{193}) and {\it MTL-Pip}(\ding{193}$\rightarrow$\ding{192}) show much better performance than {\it MTL}(\ding{192}+\ding{193}), demonstrating that our framework is able to better utilize the interdependence between subtasks than {\it MTL}.
This is due to the utilization of bi-directional interaction between these two tasks: the previous task provides useful information to the subsequent task, meanwhile, the back propagation of the subsequent task can also positively affect its previous one.
Furthermore, the performance of
{\it MTL-Pip}(\ding{192}$\rightarrow$\ding{193}) is better than {\it MTL-Pip}(\ding{193}$\rightarrow$\ding{192}). Thus, we believe that the direction of jointly modeling subtasks has an important effect on our framework.
\textbf{Third},
jointly modeling all three subtasks (the last group in Table \ref{tab:SC_result}) is better than modeling two tasks (the second last group in Table \ref{tab:SC_result}), no matter which framework ({\it  MTL-Pip} or {\it MTL}) is used.
This result suggests that any subtask can provide useful information to other subtasks.
\textbf{Finally}, 
the better performance of {\it MTL-Pip}(\ding{192}$\rightarrow$\ding{193}$\rightarrow$\ding{194}) regarding {\it MTL-Pip}(\ding{193}$\rightarrow$\ding{192}$\rightarrow$\ding{194}) confirms that it is better to stack \ding{193} upon \ding{192} (\ding{192}$\rightarrow$\ding{193}) via pipeline. Probably, this is because simile sentence classification (\ding{192}, binary classification) is generally easier than simile component extraction (\ding{193}, sequence labeling), and thus it is more reasonable to finish the easy task before the difficult one.

\subsubsection{Effect of Cyclic Multitask Learning Framework}
As shown in Table \ref{tab:SC_result} ({\bf Line 10-13}), we can see that our \emph{MTL-Cyc} outperforms other contrast systems, including \emph{MTL-Pip}(\ding{192}$\rightarrow$\ding{193}$\rightarrow$\ding{194}).
Note that \emph{MTL-Pip}(\ding{192}$\rightarrow$\ding{193}$\rightarrow$\ding{194}) is a subset of our cyclic framework and has not been investigated before.
\emph{MTL-Cyc} is better than previous numbers and our strong baselines, demonstrating its effectiveness.

\subsection{Task 2: Simile Component Extraction}

Table \ref{tab:CE_result} shows the comparison results on simile component extraction. 
Similar to the experimental results on Task 1, our \emph{MTL-Cyc} and \emph{MTL-Pip} still beats other models.
Specially, our {\it MTL-Pip}(\ding{192}$\rightarrow$\ding{193}$\rightarrow$\ding{194}) exhibits better performance than {\it MTL-OP \cite{Liu:EMNLP2018}}, showing the effectiveness of information sharing between subtasks during training.
Moreover, we discuss the results from the following aspects:

\begin{table}[t]
\centering
\small
\setlength{\tabcolsep}{1.5mm}{
\begin{tabular}{l|ccc}  
\hline
\textbf{Model}  & \textbf{Precision} & \textbf{Recall} & \textbf{F1-score} \\
\hline          
MTL \cite{Liu:EMNLP2018} & 55.99  & 69.89   & 62.11 \\
MTL-OP \cite{Liu:EMNLP2018} & 61.60  & 73.61  & 67.07 \\
\hline
\ding{193}   & 54.98  &  66.47  & 60.18 \\
\hline
MTL(\ding{192}+\ding{193})  & 55.46  & 65.09   & 59.89 \\
MTL(\ding{193}+\ding{194})  & 58.07  & 69.53   & 63.29 \\
MTL-Pip(\ding{192}$\rightarrow$\ding{193})   & 63.14  & 70.12   & 66.45 \\
MTL-Pip(\ding{193}$\rightarrow$\ding{192})   & 56.87  & 67.75   & 61.84 \\
\hline
MTL(\ding{192}+\ding{193}+\ding{194})  & 55.92  & 71.30   & 62.68 \\
MTL-Pip(\ding{192}$\rightarrow$\ding{193}$\rightarrow$\ding{194}) & \textbf{64.08}  &  71.60  & 67.63 \\
MTL-Pip(\ding{193}$\rightarrow$\ding{192}$\rightarrow$\ding{194}) & 57.54  & 73.37  & 64.50 \\
MTL-Cyc & 63.16  & \textbf{73.78} & \phantom{*}\textbf{68.05}* \\
\hline
\end{tabular}}
\caption{Main results on simile component extraction. * indicates significant at $p<0.01$ over \emph{MTL(\ding{192}+\ding{193}+\ding{194})}.}
\label{tab:CE_result}
\end{table}

\subsubsection{Effects of Simile Sentence Classification and Sentence Reconstruction}
We first investigate contributions of simile sentence classification and sentence reconstruction to simile component extraction via {\it MTL} and {\it MTL-Pip} frameworks.
Here, 
we draw the following conclusions:
\textbf{First}, 
as same as our results on Task 1,
both {\it MTL-Pip}(\ding{192}$\rightarrow$\ding{193}) and {\it MTL-Pip}(\ding{193}$\rightarrow$\ding{192}) obtain better performance than {\it MTL}(\ding{192}+\ding{193}).
It confirms the superiority of our framework, which enables the involving subtasks to better benefit from each other than the conventional multitask learning.
\textbf{Second}, 
{\it MTL}(\ding{192}+\ding{193}) shows worse results than {\it MTL}(\ding{193}+\ding{194}). This may be owing to \ding{194} (sentence reconstruction) bringing more supervision signals over the encoder than \ding{192} (binary classification).
\textbf{Third}, 
{\it MTL-Pip}(\ding{192}$\rightarrow$\ding{193}) significantly outperforms {\it MTL-Pip}(\ding{193}$\rightarrow$\ding{192}). 
This observation confirms that simile sentence classification is much easier than simile component extraction. 
Thus, 
the previous simile sentence classifier can provide useful information for the subsequent simile component extraction.

\subsubsection{Effect of Cyclic Multitask Learning Framework}
Our {\it MTL-Cyc} outperforms all \emph{MTL-Pip} models, even with the same number of parameters.
This result indicates that our cyclic setting mitigates the error propagation and truly benefits from leveraging the interdependence between subtasks.

\subsection{Analysis}
In order to better understand the individual effectiveness of our proposed local attention mechanism and cyclic multitask learning framework, we carry out additional analysis from the following aspects.

\subsubsection{Distribution of Window Size}
\begin{figure}[!t]
\centering
\includegraphics[width=1.0\linewidth]{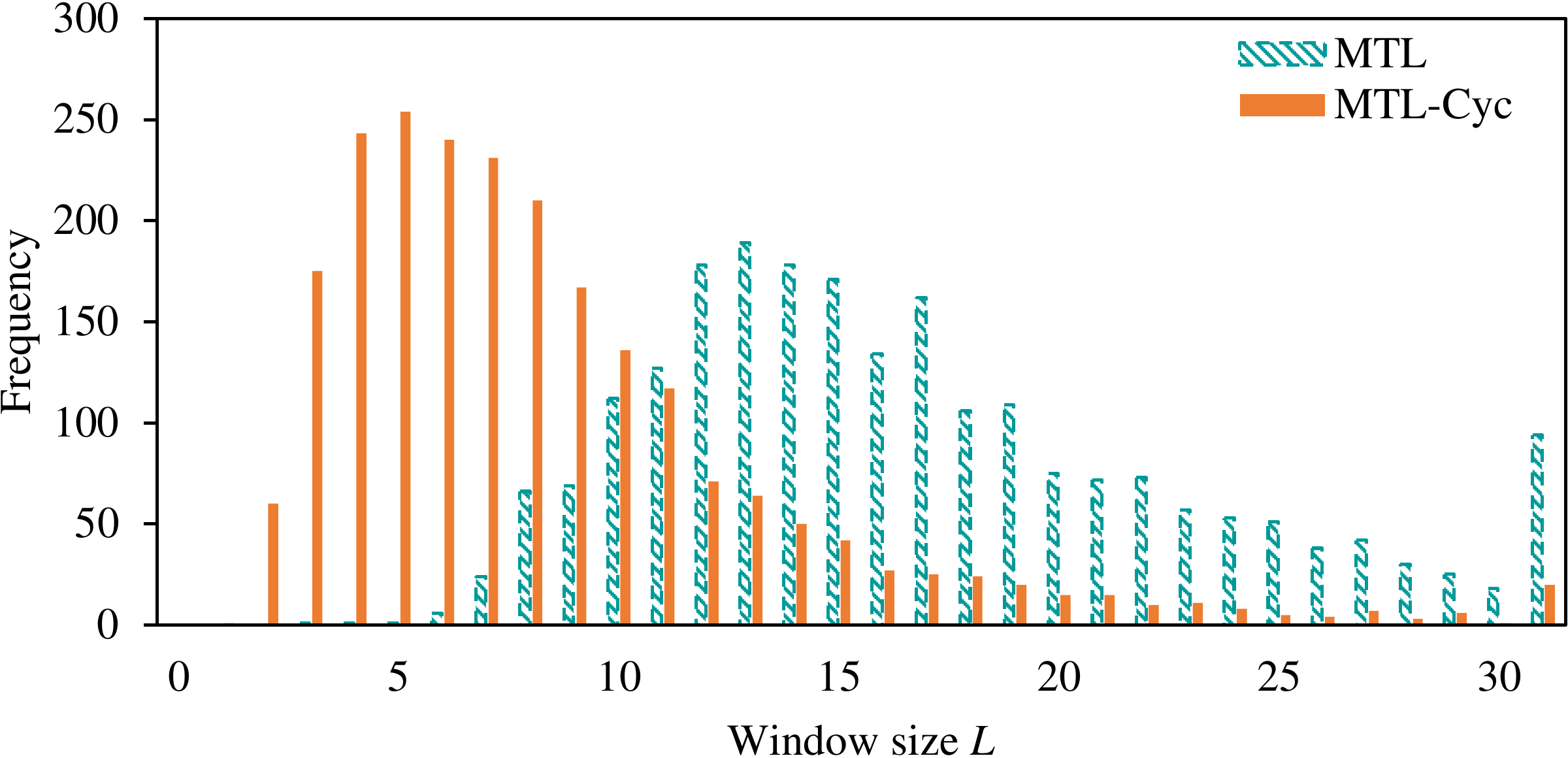}
\caption{\label{Freq_win} Distribution of the predicted window size $L$.}
\end{figure}

Figure \ref{Freq_win} shows the distribution of the predicted window size $L$ (Eq. \ref{eq:l}) for our local attention on testset.
The window sizes predicted by {\it MTL-Cyc} tend to be much smaller than those predicted by {\it MTL}.
The underlying reason is that our {\it MTL-Cyc} allows its local attention module  to be enhanced and better adjusted by the outputs and loss signals of the other tasks.

\subsubsection{F1-scores over Distances between a Tenor and a Vehicle}
\begin{figure}[!t]
\centering
\includegraphics[width=1.0\linewidth]{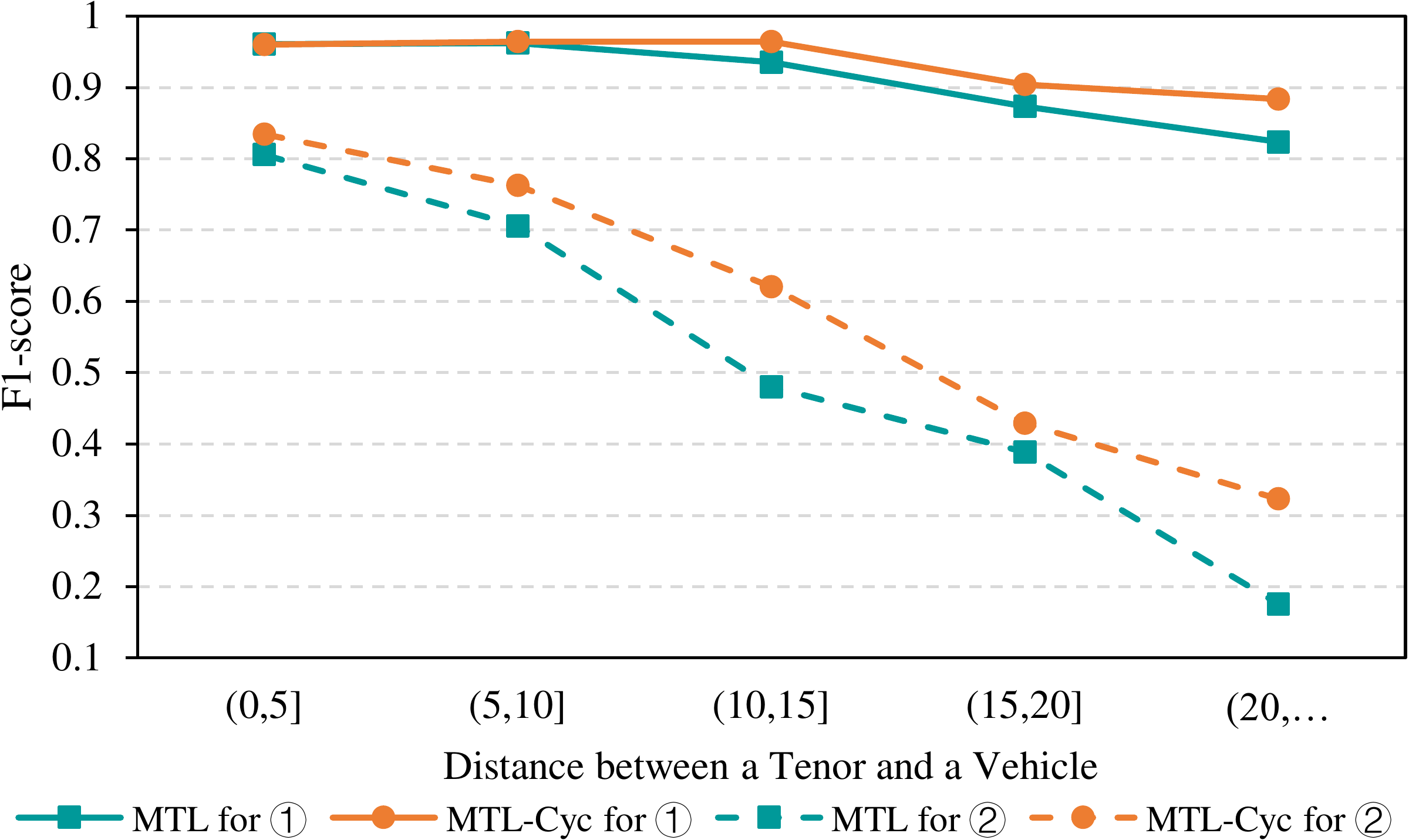}
\caption{\label{F1_DIS}
F1-scores on different groups of test instances according to the distance between a tenor and a vehicle.
Solid lines are results on simile sentence classification (\ding{192}), dashed lines are results on simile component extraction (\ding{193}).}
\end{figure}

We hypothesize that the difficulty of simile recognition increases as the distance between a tenor and a vehicle increases. 
To testify this,
we display results on different groups of test examples regarding this distance in Figure \ref{F1_DIS}.
In both tasks, we find that the advantage of our framework becomes more obvious as the distance increases.
Particularly, the performance gap between our \emph{MTL-Cyc} and \emph{MTL} becomes greater up to 15 points for difficult instances.

\subsubsection{F1-scores over Sentence Lengths}

\begin{figure}[!t]
\centering
\includegraphics[width=1.0\linewidth]{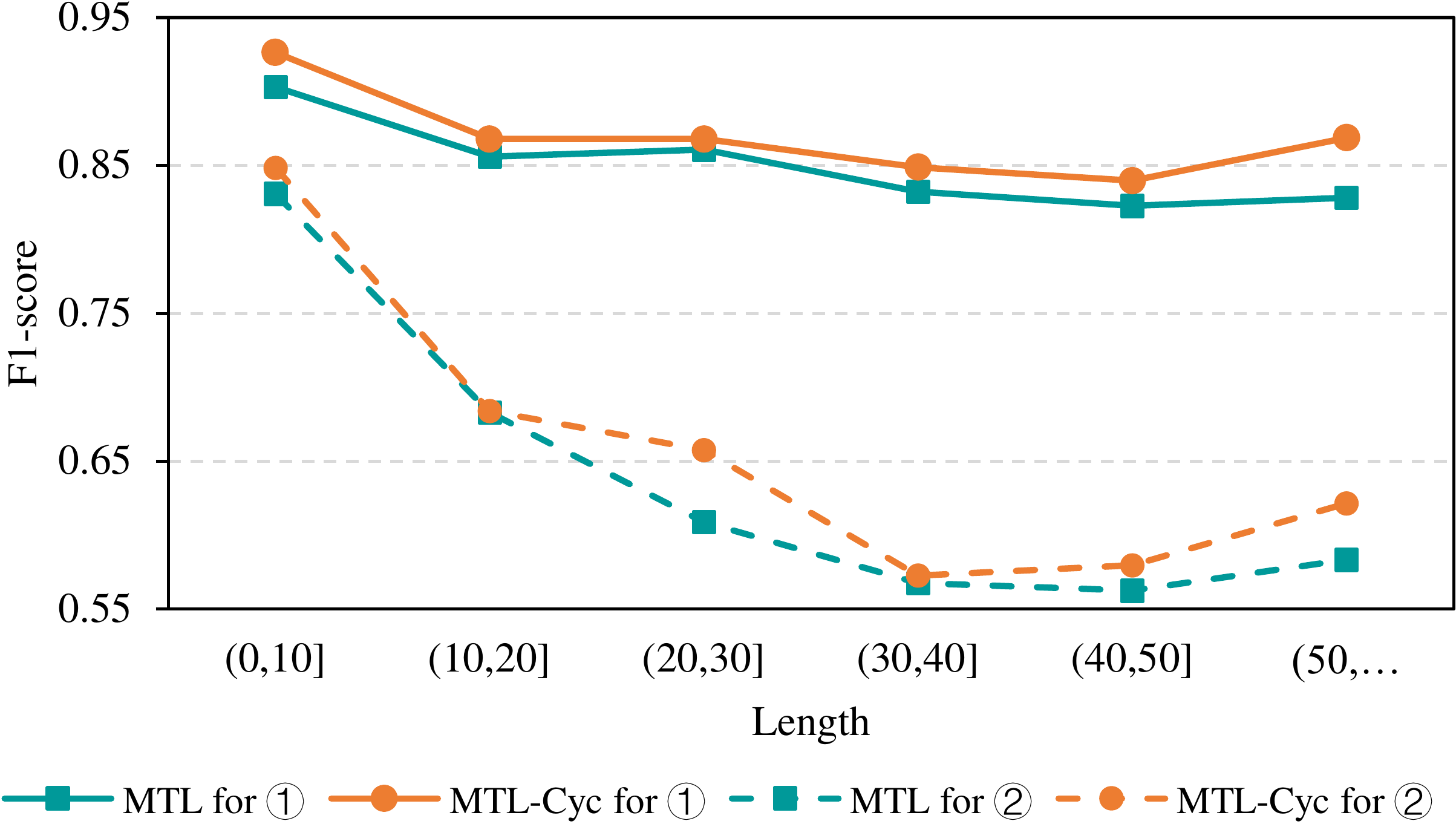}
\caption{\label{F1_Len}
F1-scores on different groups of test instances according to sentence lengths.}
\end{figure}

As shown in Figure \ref{F1_Len}, 
we compare our cyclic framework with standard MTL regarding different ranges of sentence lengths.
Results show that our framework is consistently better than \emph{MTL} in all groups, showing its robustness.

\begin{table}
	\centering
	\begin{tabular}{lccc}  
		\hline
		\textbf{Model(+BERT)}  & \textbf{Precision} & \textbf{Recall} & \textbf{F1-score} \\
		\hline
		\multicolumn{4}{l}{\textbf{Task 1: Simile Sentence Classification}} \\
		MTL(\ding{192}+\ding{193}) & 83.31  & 93.61   & 88.16 \\
		MTL-Pip(\ding{192}$\rightarrow$\ding{193}) & 83.84  &  \textbf{94.63}  & 88.91 \\
		MTL-Cyc   & \textbf{85.81}  & 94.43  & \textbf{89.92} \\
		\hline
		\multicolumn{4}{l}{\textbf{Task 2: Simile Component Extraction}} \\
		MTL(\ding{192}+\ding{193}) & 71.95  & 74.65   & 73.28 \\
		MTL-Pip(\ding{192}$\rightarrow$\ding{193}) & 72.17  &  \textbf{77.81}  & 74.88 \\
		MTL-Cyc   & \textbf{73.97}  & 77.61  & \textbf{75.74} \\
		\hline
	\end{tabular}
	\caption{Test results for simile recognition and extraction using pretrained BERT.}
	\label{tab:BERT_result}
	\vspace{-1.0em}
\end{table}

\subsubsection{Effect using BERT \cite{Devlin:NAACL2019}}
Recently, BERT has achieved great success in many NLP tasks by leveraging the rich knowledge within large-scale raw text via pretraining.
To further demonstrate the effectiveness of our framework, we replace the pretrained word embeddings with the outputs of a Chinese BERT model\footnote{https://github.com/ymcui/Chinese-BERT-wwm}, which is finetuned during simile training.
Since the Bi-LSTM sentence decoder module (step \ding{194} in Figure \ref{OurModel}) works similarly with BERT by providing additional language modeling loss,
we remove this module and directly construct a cycle with the other two tasks by feeding the outputs of simile component extraction (\ding{193} in Figure \ref{OurModel}) as additional inputs to simile sentence classification (\ding{192} in Figure \ref{OurModel}).

As shown in Table \ref{tab:BERT_result}, our \emph{MTL-Cyc} is still significantly better than \emph{MTL} and \emph{MTL-Pip} in both two subtasks given a strong pretrained BERT.
Especially, the gains over \emph{MTL} are almost 2.0 and 2.5 absolute points for simile sentence classification and simile component extraction, respectively.
Both results confirm the superiority of our \emph{MTL-Cyc} framework over standard MTL and other alternatives.

\section{Related Work}
\subsubsection{Simile Recognition and its Applications}
Similes have been studied in linguistics and psycholinguistics to explore how humans process similes, comparisons, metaphors, and the interplay among different components of these linguistic forms. 
Recently, simile recognition has wide applications in many tasks.
Veale and Hao~\shortcite{Veale-Hao:AMCSS2007} and Veale~\shortcite{Veale:MIUCC2012} showed that the category-specific knowledge acquired from explicit similes can help to better understand figurative languages, such as metaphor and irony.
Qadir et al.~\shortcite{Qadir-Riloff-Walker:EMNLP2015} studied simile on sentiment classification, because people sometimes use simile to express their feelings instead of sentiment words.
Since simile is very beneficial to other applications, simile recognition has received increasing interests in industrial and academic research.
Li et al.~\shortcite{Li:CIP2008} introduced a maximum entropy model as simile sentence classifier and a CRF as simile component extractor.
Niculae and Yaneva~\shortcite{Niculae-Yaneva:ACL2013} and Niculae~\shortcite{Niculae:JSSP2013} recognized comparisons and similes through the use of syntactic patterns. 
Niculae and Danescu-Niculescu-Mizil~\shortcite{Niculae-Mizil:EMNLP2014} distinguished simile in product reviews using a series of linguistic cues as features. 
Overall, these approaches were primarily based on handcrafted linguistic features and syntactic patterns.
Inspired by successful applications of multitask learning,
Liu et al.~\shortcite{Liu:EMNLP2018} introduced a neural multitask learning framework. We are in line with Liu et al.~\shortcite{Liu:EMNLP2018} in multitask modeling, but are different in that we consider the intercorrelation between different subtasks of simile recognition.

\subsubsection{Multitask Learning}
Recently, joint modeling multiple closely related tasks with shared representations has achieved 
great success on many NLP tasks, such as parsing and named entity recognition (NER) \cite{Finkel:ACL2010}, NER and linking \cite{Luo:EMNLP2015}, text classification \cite{Liu:IJCAI2016}, POS tagging and parsing \cite{Zhang:ACL2016}, extraction of entities and
relations \cite{Miwa:ACL2016}, event detection and summarization \cite{Wang:IJCAI2017} and simile recognition \cite{Liu:EMNLP2018}. 
Unlike previous work, 
our framework further considers the interactions between subtasks by cyclic information propagation.
On the simile recognition task, whose subtasks have strong intercorrelation between each other, our framework shows much stronger performance than conventional multitask learning framework.

\subsubsection{Local Attention}
In addition to neural machine translation \cite{Luong:EMNLP2015,Yang:EMNLP2018}, local attention mechanism has been shown effective on other NLP tasks, such as natural language inference \cite{Sperber-Niehues-Neubig-Stuker-Waibel:Interspeech2018}.
We are the first to investigate the local attention mechanism on simile recognition.

\subsubsection{Simile Recognition vs. Aspect-level Sentiment Analysis}
The task of simile recognition looks similar to aspect-level sentiment classification (ASC) \cite{Liu:SLHLT2012,Pontiki:COLING2014}.
ASC is to determine the sentiment regarding a certain aspect, while simile recognition is to detect whether there is a simile regarding a comparator word in a sentence \cite{Liu:EMNLP2018}.
However, simile recognition also requires extracting the corresponding \emph{tenor} and \emph{vehicle} if there is a simile sentence, while ASC does not require extracting any supporting evidence from text.
Hence, the existing work for ASC can not be simply applied for simile recognition with a naive adaptation.
More importantly,
to the best of our knowledge, our framework is also novel and has large potential in the field of ASC.
Specifically,
the current state-of-the-art ASC model \cite{Hu:ACL2018} shows that a decoding pipeline, which explicitly explores the intercorrelation among subtasks, surprisingly outperforms standard MTL \cite{Li:AAAI2019}, even though decoding pipelines generally suffer from error propagation.
This indicates that our cyclic-MTL may further improve ASC, as the joint training and cyclic flow of our framework can better model the intercorrelation among subtasks than a simple decoding pipeline.
We leave studying our cyclic-MTL on ASC for future work.


\section{Conclusion}

We presented a novel cyclic multitask learning framework 
for simile recognition.
Compared with conventional multitask learning, our framework can better model the dependencies among the subtasks.
Extensive experiments and 
analysis strongly demonstrate the effectiveness of our framework.

In the future,
we plan to investigate the generality of our framework on 
other multitask learning based NLP tasks.
Besides, 
we will explore how to improve our framework by introducing variational networks,
which have been widely used in many tasks \cite{Zhang:EMNLP2016b,Zhang:EMNLP2016a,Su:AAAI2018,Su:INS2018}. 

\section{Acknowledgments}
The authors were supported by Beijing Advanced Innovation Center for Language Resources, National Natural Science Foundation of China (No. 61672440), the Fundamental Research Funds for the Central Universities (Grant No. ZK1024), and Scientific Research Project of National Language Committee of China (Grant No. YB135-49). 

\bibliographystyle{aaai}
\bibliography{aaai}

\end{document}